\documentclass[11pt]{article}

\usepackage[final]{acl}

\usepackage{times}
\usepackage{latexsym}
\usepackage{outlines}
\usepackage{amsmath}
\usepackage{booktabs}
\usepackage{placeins}

\usepackage[T1]{fontenc}

\usepackage[utf8]{inputenc}

\usepackage{microtype}

\usepackage{inconsolata}

\usepackage{graphicx}

\title{LLM-as-a-Judge for Low-Resource Languages: Adapting Ragas and Comparative Ranking for Romanian}

\author{
Claudiu Creanga$^{2, 3}$, Liviu P. Dinu$^{1, 3}$\\ 
  $^1$ Faculty of Mathematics and Computer Science, \\
  $^2$ Interdisciplinary School of Doctoral Studies, \\
  $^3$ HLT Research Center, \\
  University of Bucharest, Romania\\
  \small
{\tt claudiu.creanga@fmi.unibuc.ro, ldinu@fmi.unibuc.ro}  \\
}

\begin{document}
\maketitle
\begin{abstract}
Evaluating Retrieval-Augmented Generation (RAG) systems remains a challenge for Low-Resource Languages (LRLs), where standard reference-based metrics fall short. This paper investigates the viability of the "LLM-as-a-Judge" paradigm for Romanian by adapting the Ragas framework using next-generation models (Gemini 2.5 and Gemini 3). We introduce \textbf{AdminRo-Eval}, a curated dataset of Romanian administrative documents annotated by native speakers, to serve as a ground truth for benchmarking automated evaluators. We compare three evaluation methodologies—direct scoring, comparative ranking, and granular decomposition—across metrics for Faithfulness, Answer Relevance, and Context Relevance. Our findings reveal that evaluation strategies must be metric-specific: granular decomposition achieves the highest human alignment for \textit{Faithfulness} (96\% with Gemini 2.5 Pro), while comparative ranking outperforms in \textit{Answer Relevance} (90\%). Furthermore, we demonstrate that while lightweight models struggle with complex reasoning in LRLs, the Gemini 2.5 Pro architecture establishes a robust, transferable baseline for automated Romanian RAG evaluation.
\end{abstract}

\section{Introduction}

Building AI systems is inherently an iterative loop of design, evaluation, and refinement. However, in the context of \textbf{Low-Resource Languages (LRLs)}, this loop is often disrupted by a scarcity of reliable benchmarks \cite{JoshiEtAl2020}. While recent initiatives have started to bridge this gap for Romanian in tasks like lexical simplification \cite{anghel2025rals}, resources for complex reasoning pipelines remain limited. In the era of LLMs, where outputs are non-deterministic, the ability to rapidly and accurately measure system fidelity is important. Evaluation in this context serves as the functional equivalent of unit tests in traditional software engineering, providing the essential signal required to distinguish improvement from regression, a signal that is frequently absent or noisy for under-represented languages \cite{KreutzerEtAl2022}.

To address the intrinsic limitations of frozen LLMs, specifically their tendency toward \textbf{hallucinations} and their inability to access information beyond their training data cutoffs, RAG systems have emerged as the dominant architectural paradigm. By dynamically anchoring the model's generative capabilities to verifiable, external knowledge sources, RAG reduces hallucinations. While the conceptual framework of RAG is well-established, evaluating these pipelines presents unique challenges for the low-resource community.

A significant contribution in this domain is the \textbf{Ragas framework} \cite{es2023ragas}, which introduced a suite of reference-free metrics (such as context precision, faithfulness, and answer relevance) for the automated assessment of RAG pipelines. This framework leverages an "LLM-as-a-Judge" approach, utilizing an advanced model to critique the performance of the RAG system. However, the original validation of Ragas was heavily English-centric and relied on the now-dated \textbf{gpt-3.5-turbo} architecture \cite{ouyang2022training}. This English focus leaves a gap regarding the framework's applicability to lower-resource languages like \textbf{Romanian}, which often suffer from performance degradation when using tools optimized for English. We contend that for an "LLM-as-a-Judge" to be effective in a low-resource setting, the judge itself must possess state-of-the-art multilingual reasoning capabilities to compensate for the lack of training data. Consequently, we focus on applying the Ragas framework to Romanian datasets by replacing legacy evaluators with next-generation engines: \textbf{Gemini 2.5} \cite{geminiteam2023gemini} and \textbf{Gemini 3} \cite{geminiteam2025gemini3}. By leveraging these newer, more powerful models, we aim to establish a robust baseline for automated RAG evaluation in Romanian, providing a transferable strategy for other low-resource languages.

Our \textbf{contributions} are:

\begin{itemize}
    \item \textbf{First Application of Ragas to Romanian}: We conduct the first rigorous evaluation of the RAGAs framework in a non-English context. By validating these metrics on Romanian data, we demonstrate their cross-lingual portability and establish a precedent for reference-free evaluation in LRLs.
    
    \item \textbf{Benchmarking Next-Generation Evaluators}: Deviating from the framework's original reliance on GPT-3.5, we integrate and benchmark a suite of modern architectures, including \textbf{Gemini 2.5 Pro}, \textbf{Gemini 3 Pro}, and \textbf{Gemini 3 Flash}. Our comparative analysis reveals that specific Gemini variants significantly outperform legacy baselines in aligning automated scores with human judgment, challenging the "one-size-fits-all" approach to model selection.

    \item \textbf{Introduction of AdminRo-Eval}: We release a novel, domain-specific evaluation dataset curated from official Romanian public administration documents. Named \textbf{AdminRo-Eval}, this resource provides a challenging, real-world benchmark for RAG systems in a formal, knowledge-intensive domain. It is specifically designed to aid in the development of public sector AI agents, offering a direct metric for how accurately systems can retrieve and synthesize official documentation for citizen queries.
\end{itemize}

\section{Related Work}

Evaluating RAG systems is uniquely challenging in low-resource settings where "gold standard" reference texts are virtually non-existent. Historically, metrics like BLEU \cite{papineni-etal-2002-bleu} and ROUGE \cite{lin-2004-rouge} have dominated NLP. However, their reliance on lexical overlap with human-curated references makes them inadequate for RAG, where correct answers often diverge in vocabulary. For Low-Resource Languages (LRLs), where creating high-quality reference datasets is prohibitively expensive, these metrics are particularly ill-suited. Furthermore, while detecting the \textit{presence} of machine-generated text is feasible \cite{marchitan2024transformer}, assessing its \textit{accuracy} without gold references is far more complex.

While embedding-based metrics like BERTScore \cite{zhang2020bertscore} improve upon lexical rigidity, they often typically retain the dependency on reference text. To bypass this data bottleneck, the field has shifted toward the "LLM-as-a-Judge" paradigm \cite{zheng2023judging}. This approach employs a strong LLM as an adjudicator, offering a scalable alternative to human evaluation.

External verification methods like FActScore \cite{min2023factscore}, which cross-reference claims against knowledge bases (e.g., Wikipedia), face scalability hurdles in low-resource domains where such bases are often sparse or outdated. Consequently, the \textbf{Ragas framework} \cite{es2023ragas} has emerged as a vital tool. By atomizing evaluation into reference-free metrics such as \textbf{Faithfulness} and \textbf{Context Relevance}, Ragas assesses reasoning directly. Our work adopts this methodology to determine if such reference-free frameworks can be robustly transferred to the Romanian linguistic context.

\begin{figure*}[htbp]
    \centering
    \includegraphics[width=0.8\linewidth]{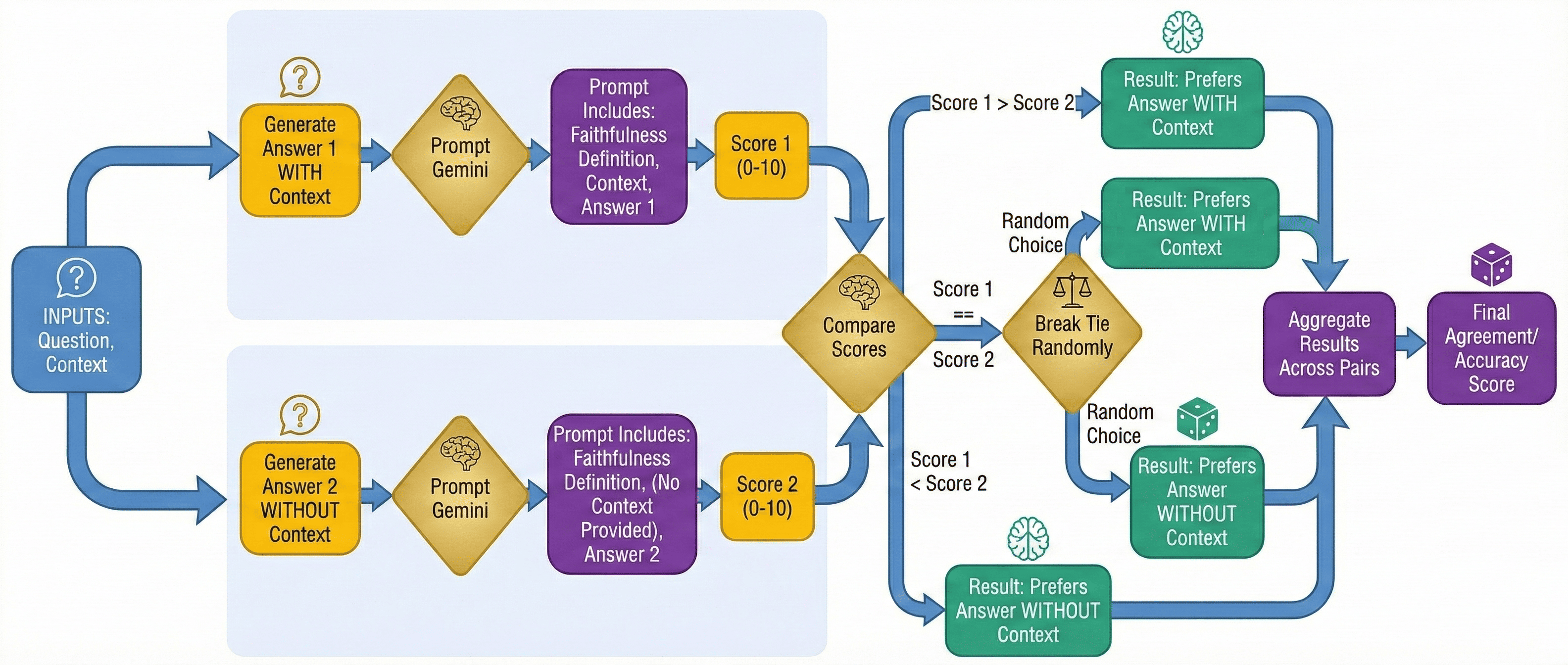} 
    \caption{Operational flow of the Baseline (LLM-as-Judge) evaluation. This method isolates the assessment process by prompting the LLM to assign an independent scalar score (0-10) to the output.}
    \label{fig:faithfulness_baseline}
\end{figure*}

\begin{figure*}[htbp]
    \centering
    \includegraphics[width=0.8\linewidth]{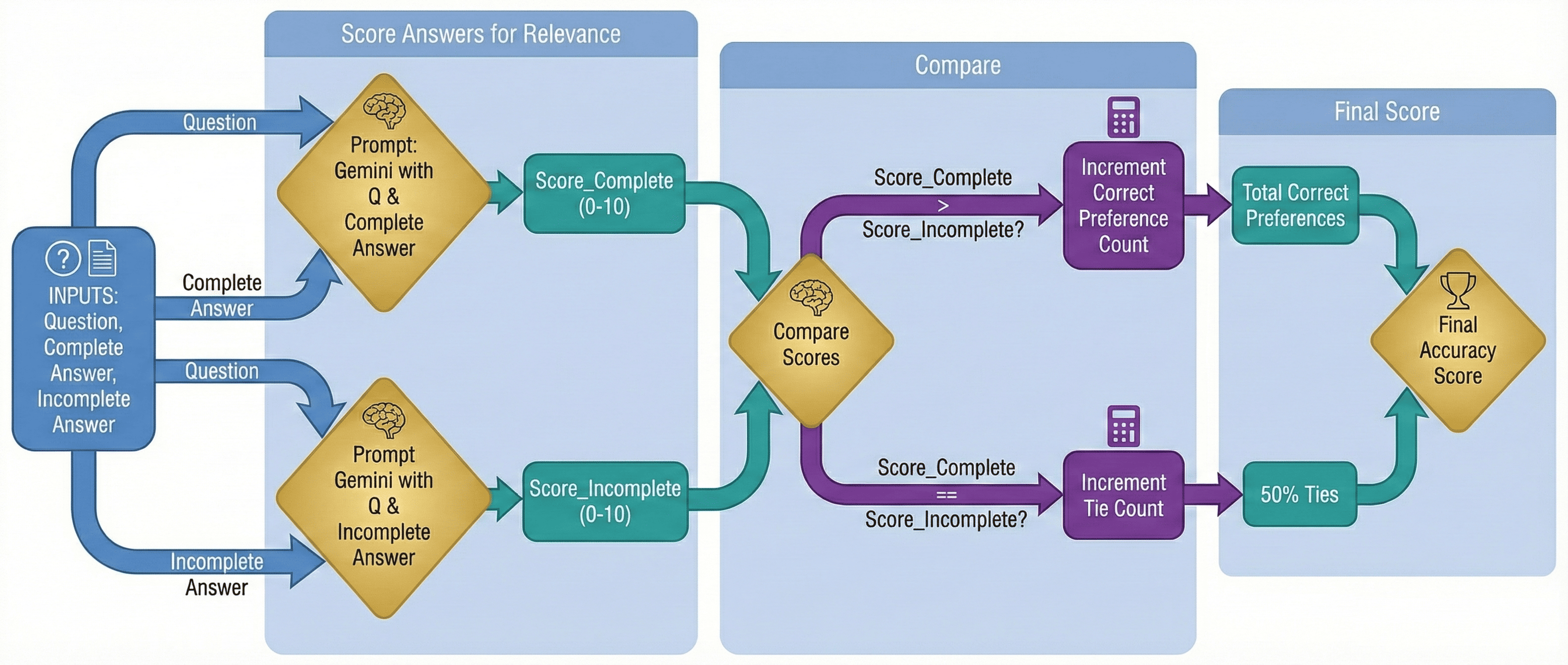} 
    \caption{\textbf{Answer Relevance Baseline.} Gemini independently scores a complete and an incomplete answer. The scores are compared to determine correct preferences.}
    \label{fig:answer_baseline}
\end{figure*}

\begin{figure*}[htbp]
    \centering
    \includegraphics[width=0.8\linewidth]{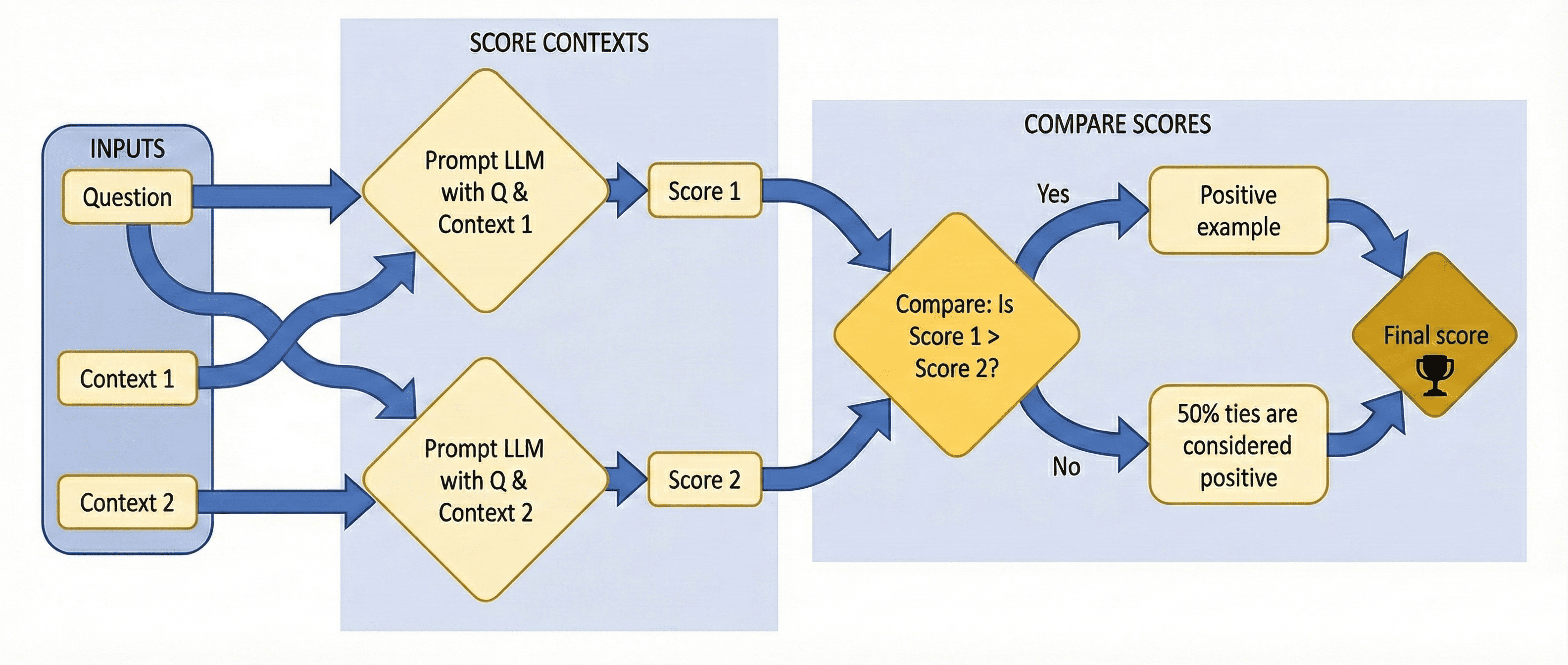} 
    \caption{\textbf{Context Relevance Baseline.} The LLM independently assigns scalar scores to a relevant and an irrelevant context relative to the query.}
    \label{fig:context_baseline}
\end{figure*}

\section{Methodology}

\subsection{Data Source}

Official documents from Romanian public administration bodies provided the raw material for our corpus. The pipeline for gathering and preparing this data involved the following steps:

\begin{itemize} 
\item \textbf{Data Sources}: The initial collection comprised 84 PDF documents from institutions such as the Romanian Health Ministry, the Senate, and official publishing houses. These sources guarantee that the dataset contains the formal language and complex subject matter typical of real-world retrieval tasks. 
\item \textbf{Filtering}: We restricted the dataset to documents published within the two months preceding collection. This temporal filter prevents data leakage, ensuring the LLM did not encounter the text during its training. We further refined the selection by removing pages containing mostly figures, signatures, or blank space, leaving 27 usable multi-page PDFs. 
\item \textbf{Chunking}: Since RAG systems process smaller segments rather than full documents, we divided the filtered PDFs into contiguous 3-page blocks. This process yielded a final set of 100 specific PDF chunks to serve as the knowledge context. \end{itemize}

\subsection{Data Generation in Romanian}

Creating clear test cases required generating contrastive pairs for each metric: one correct example and one intentionally flawed counterpart. This binary approach simplifies the decision for both human judges and the LLM. We conducted all prompting and generation using the Gemini model in Romanian to maintain linguistic accuracy.

For every one of the 100 PDF chunks, the process included:

\begin{outline} 
\1 \textbf{Question Extraction}: Using prompts in Romanian, Gemini analyzed each PDF chunk to formulate a single, text-supported question. This resulted in 100 base questions. 
\1 \textbf{Generating Paired Data for Evaluation}: 
\2 \textbf{For Faithfulness}: We aimed to distinguish between grounded answers and hallucinations. First, the model generated "Answer 1" using the provided context. We then requested a response to the same question without supplying the context, forcing the model to rely on internal memory and increasing the risk of error ("Answer 2"). 
\2 \textbf{For Answer Relevance}: We contrasted helpful responses with unhelpful ones. The model first produced a direct, complete response ("Answer 1"). A subsequent prompt explicitly requested an evasive or partial reply to the same question, creating "Answer 2." 
\2 \textbf{For Context Relevance:} The original 3-page chunk served as the precise "Context 1." To test noise handling, we added an unrelated page from a different document to the original chunk, resulting in a less relevant "Context 2." 
\end{outline}

This approach produced a dataset containing 100 contrastive pairs per metric, enabling a controlled comparison in Romanian.

\subsection{Human Annotation}

To define a gold standard for the Romanian language, we first used Gemini to generate 100 questions (one per PDF chunk) and contrastive synthetic pairs for each metric: grounded vs. ungrounded answers for Faithfulness, complete vs. evasive answers for Answer Relevance, and original vs. noisy chunks for Context Relevance. 

Two native speakers then acted as annotators, evaluating these 100 generated questions and their corresponding synthetic pairs through a comparative process.

The evaluation relied on a forced-choice format. For metrics like Faithfulness, annotators reviewed the question, the context, and two distinct answers, selecting the one that adhered better to the text (see Figure \ref{fig:annotation_faithfulness}). Similar comparative tasks were used for Context Relevance (Figure \ref{fig:annotation_context}) and Answer Relevance (Figure \ref{fig:annotation_answer}). We provided detailed guidelines in Romanian to ensure consistent scoring. These human judgments function as the ground truth for testing our automated evaluation.

\section{Framework for Romanian RAG Evaluation}

To rigorously assess the output quality of our Retrieval-Augmented Generation (RAG) system, we implemented an evaluation protocol strictly aligned with the metrics established in the Ragas research.

\subsection{Ragas Metrics}
We adapted three core metrics from the Ragas framework to specifically measure the fidelity and utility of generated Romanian content:
\begin{itemize}
    \item Faithfulness: This metric evaluates the extent to which the generated response, $a_s(q)$, is factually grounded in the retrieved context, c(q). A response is deemed faithful only if every assertion it contains can be logically inferred from the source material.
    \item Answer Relevance: This indicator assesses the alignment between the generated response and the user's initial inquiry. 
    \item Context Relevance: This metric audits the retrieval component itself. A context c(q) is classified as relevant if it provides exactly the information required to address the query, minimizing the signal-to-noise ratio by excluding superfluous content.
\end{itemize}

\section{Automated Assessment Strategies}
We conducted a comparative study of three distinct automated evaluation methodologies, utilizing the Gemini models as the adjudicators. The objective was to identify which scoring technique demonstrated the highest correlation with human-annotated ground truth.

\subsection{Method 1: Direct Scoring (LLM-as-Judge)}

This baseline approach utilizes the LLM as a scalar scorer. For every answer pair, we issue distinct prompts to the model, requesting a numerical rating between 0 and 10 based on the specific metric definitions. 

As illustrated in Figure \ref{fig:faithfulness_baseline} and Table \ref{tab:evaluation_results}, this technique secured an 86\% alignment with human judgments regarding Faithfulness when using the Gemini 2.5 Pro model and 82\% when using Gemini 3 Pro. We similarly applied this independent scoring methodology to Answer Relevance (Figure \ref{fig:answer_baseline}) and Context Relevance (Figure \ref{fig:context_baseline}).

\subsection{Method 2: Comparative Ranking (LLM-as-Ranker)}

By shifting the paradigm from absolute scoring to relative ranking, this method simplifies the evaluation task. We present Gemini with the query alongside two competing answers in a single prompt, requesting a binary preference for the superior response. 

\begin{figure*}[htbp]
    \centering
    \includegraphics[width=0.8\linewidth]{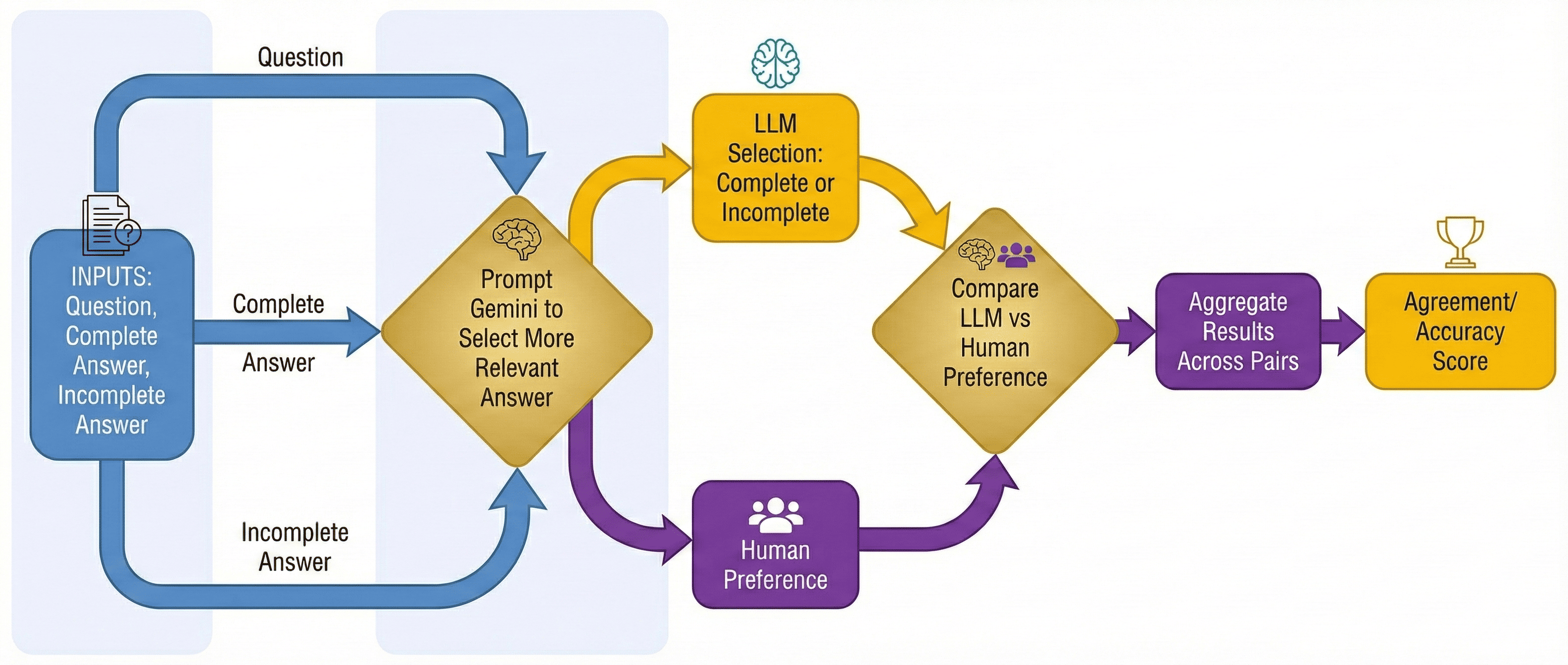} 
    \caption{\textbf{Answer Relevance Ranking}: Pairwise comparison method where Gemini is prompted to identify the more relevant response between a complete and an incomplete answer.}
    \label{fig:answer_ranking}
\end{figure*}

\begin{figure*}[htbp]
    \centering
    \includegraphics[width=0.8\linewidth]{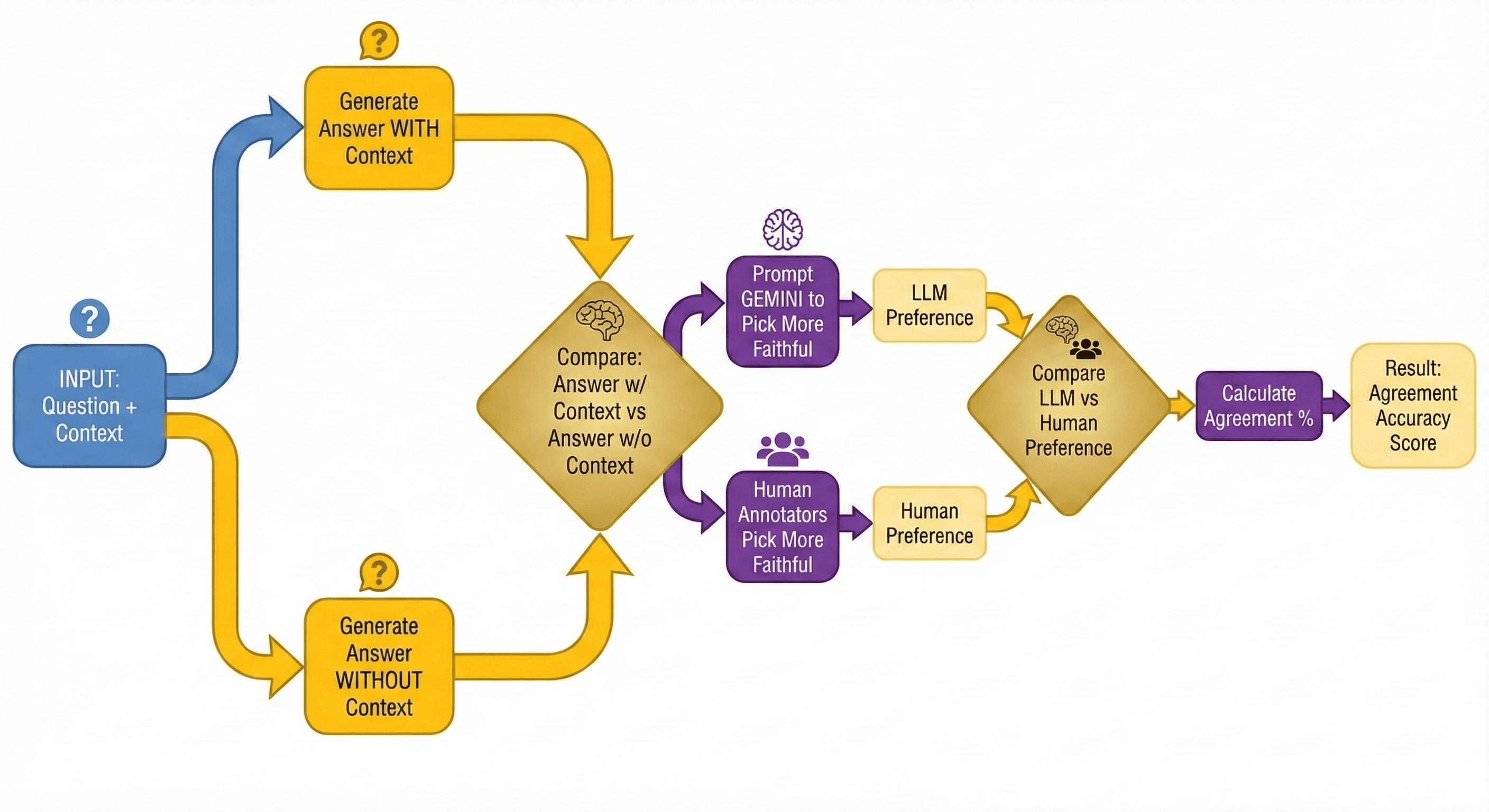} 
    \caption{\textbf{Faithfulness Ranking}: The model is presented with a simultaneous view of competing answers and must discriminatively select the candidate with higher factual grounding.}
    \label{fig:faithfulness_ranking}
\end{figure*}

\begin{figure*}[htbp]
    \centering
    \includegraphics[width=0.8\linewidth]{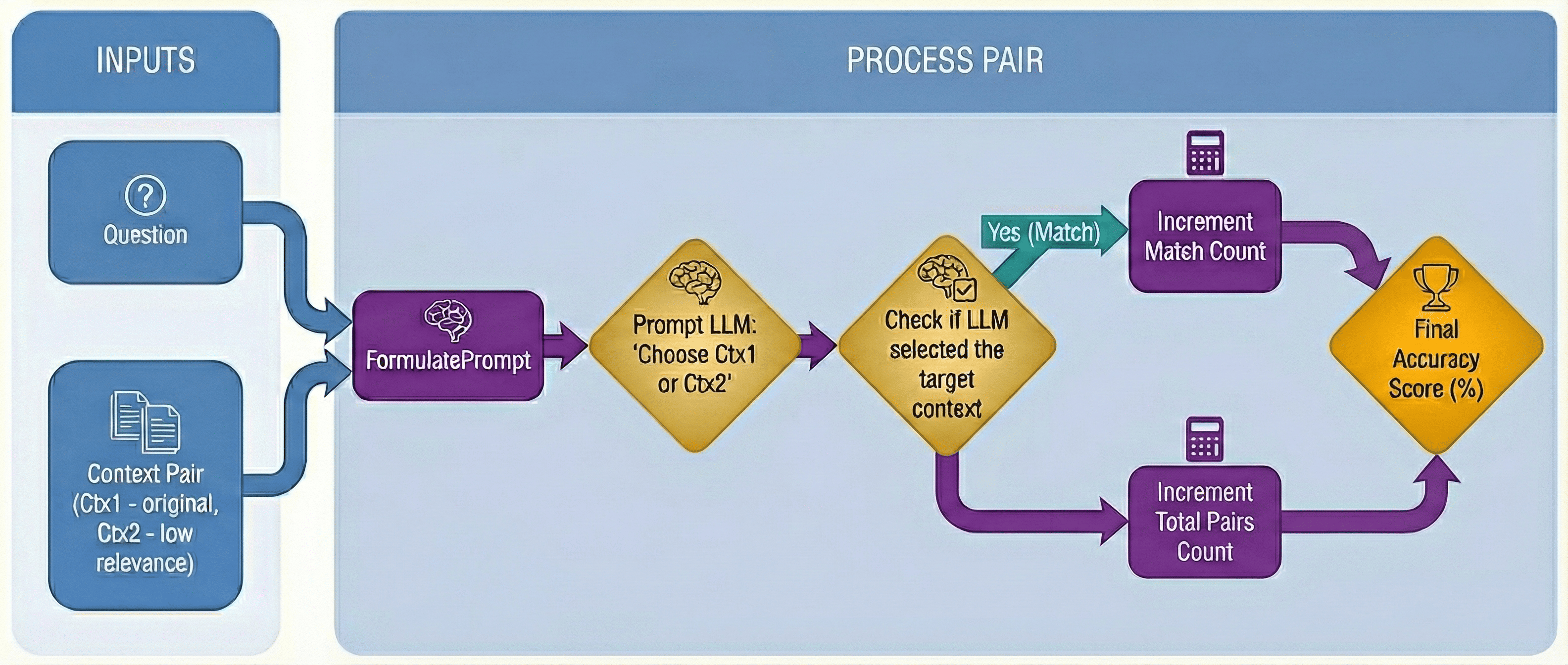} 
\caption{\textbf{Context Relevance Ranking}: Instead of independent scoring, the LLM evaluates a pair of contexts simultaneously to select the most relevant one.}
    \label{fig:context_ranking}
\end{figure*}

This pairwise comparison yielded a 90\% agreement rate with human judgments for Answer Relevance using Gemini 2.5 Pro (see Figure \ref{fig:answer_ranking} and Table \ref{tab:evaluation_results}). The corresponding ranking architectures for Faithfulness and Context Relevance are detailed in Figures \ref{fig:faithfulness_ranking} and \ref{fig:context_ranking}, respectively.

\subsection{Method 3: Granular Decomposition (Ragas-style)}

Representing the central methodology of the Ragas framework, this approach dissects the evaluation process into atomic verification steps.

\begin{figure*}[htbp]
    \centering
    \includegraphics[width=0.8\linewidth]{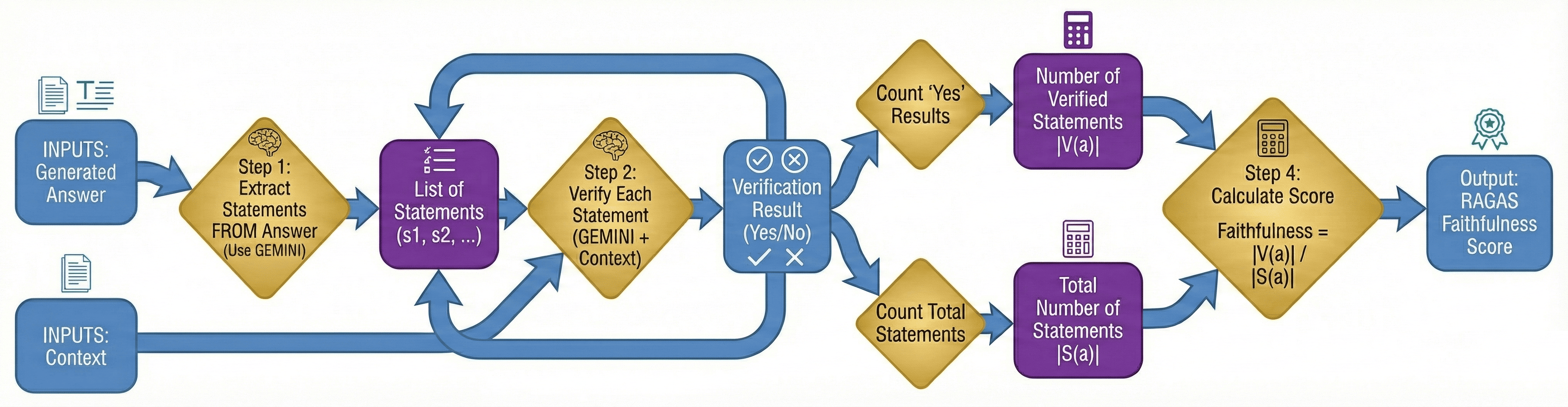} 
    \caption{\textbf{Faithfulness Decomposition}: The answer is deconstructed into atomic statements and validated against the source context.}
    \label{fig:faithfulness_ragas}
\end{figure*}

\begin{figure*}[htbp]
    \centering
    \includegraphics[width=0.8\linewidth]{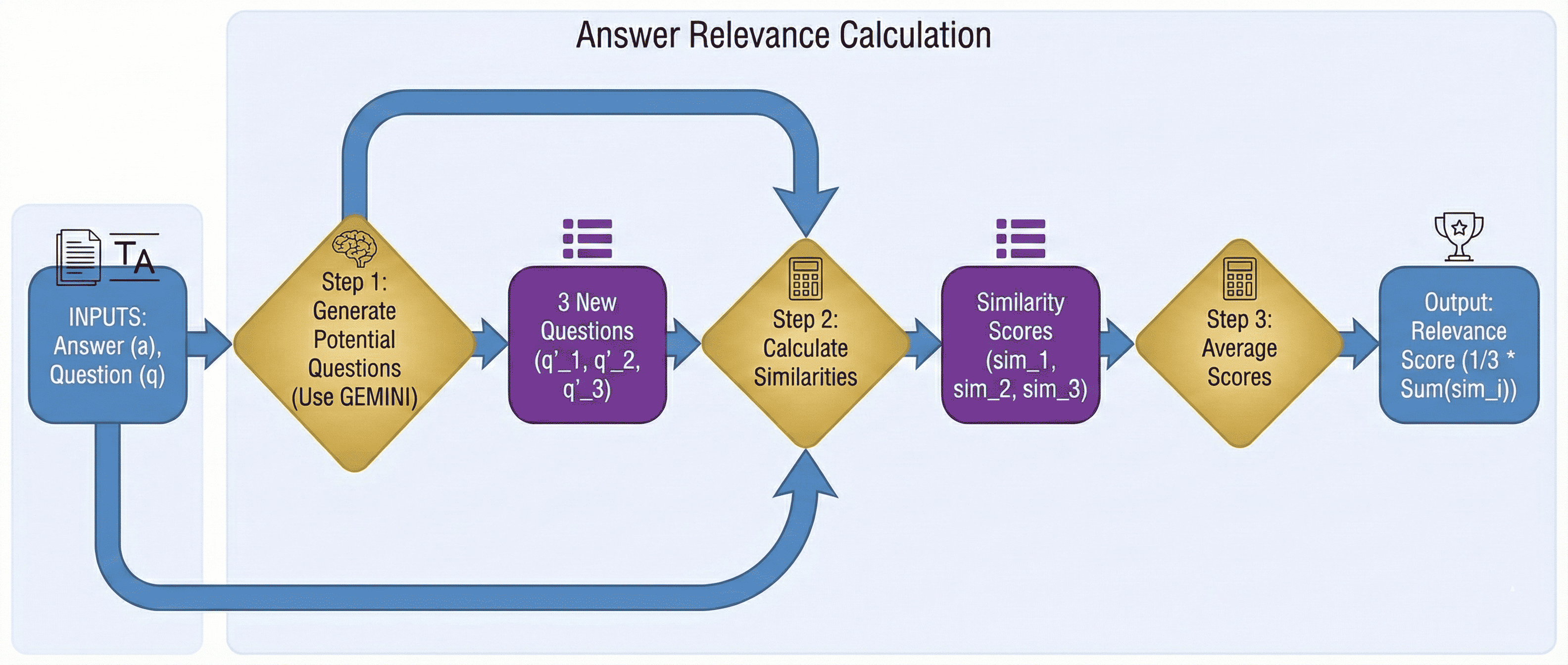} 
    \caption{\textbf{Answer Relevance Ragas}: The LLM generates hypothetical questions ($q'_i$) based on the answer to check semantic alignment with the original query.}
    \label{fig:answer_ragas}
\end{figure*}

\begin{figure*}[htbp]
    \centering
    \includegraphics[width=0.8\linewidth]{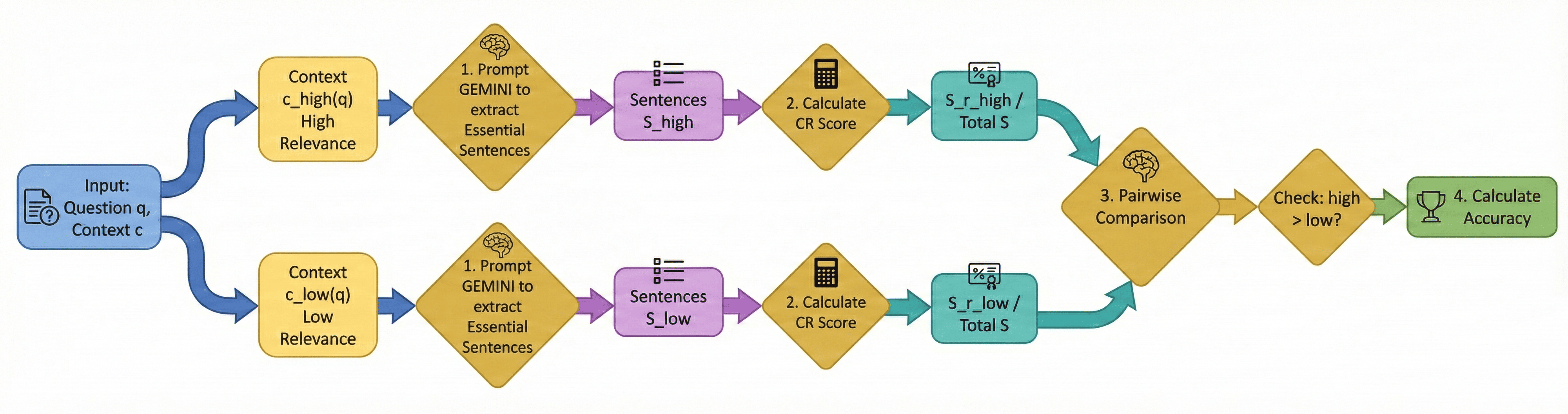} 
\caption{\textbf{Context Relevance Ragas}: The LLM extracts "essential sentences" to determine the signal-to-noise ratio of the retrieved context.}
    \label{fig:context_ragas}
\end{figure*}

\begin{itemize}
\item \textbf{Faithfulness Evaluation (Ragas-style):} As shown in Figure \ref{fig:faithfulness_ragas}, Gemini parses the Romanian response into distinct statements ($s_i$) and verifies if each is supported by the context. The final score is the ratio of verified statements to total statements:
   $$F = \frac{|V(a)|}{|S(a)|}$$
   This granular approach achieved a \textbf{96\%} agreement with human annotations when using Gemini 2.5 Pro.

\item \textbf{Answer Relevance Evaluation (Ragas-style):} 
   We prompt Gemini to synthesize $n$ potential inquiries for the generated answer (Figure \ref{fig:answer_ragas}). We then compute the cosine similarity between the original query $q$ and these synthetic questions $q'_i$:
   $$AR = \frac{1}{n} \sum_{i=1}^{n} \text{sim}(q, q'_i)$$
   This technique resulted in an \textbf{85\%} agreement for Gemini 2.5 Pro.

\item \textbf{Context Relevance (Ragas-style):}
   Illustrated in Figure \ref{fig:context_ragas}, this method tasks the model with extracting only the "essential sentences" required to answer the query. The score is calculated as the ratio of essential sentences to the total sentence count in the chunk. Here Gemini 3 pro obtained the best score of 93\%.
\end{itemize}

\begin{table*}[htbp]
    \centering
    \caption{Comparative Evaluation of LLM Judges on Romanian and English Datasets}
    \label{tab:evaluation_results}
    \begin{tabular}{lccc}
        \toprule
        \textbf{Model} & \textbf{Faithfulness} & \textbf{Answer Relevancy} & \textbf{Context Relevancy} \\
        \midrule
        \multicolumn{4}{l}{\textit{\textbf{Romanian Dataset Evaluation}}} \\
        \midrule
        Gemini 3 Pro & \shortstack[l]{\textbf{Ragas(84\%)} \\ Baseline(82\%) \\ Ranking(81\%)} & \shortstack[l]{Ragas(76\%) \\ \textbf{Baseline(83\%)} \\ Ranking(80\%)} & \shortstack[l]{\textbf{Ragas(93\%)} \\ Baseline(0\%) \\ Ranking(8\%)} \\
        \midrule
        Gemini 3 Flash & \shortstack[l]{\textbf{Ragas(87\%)} \\ Baseline(76\%) \\ Ranking(76\%)} & \shortstack[l]{Ragas(76\%) \\ Baseline(76\%) \\ \textbf{Ranking(77\%)}} & \shortstack[l]{Ragas(0\%) \\ Baseline(2\%) \\ \textbf{Ranking(5\%)}} \\
        \midrule
        Gemini 2.5 Pro & \shortstack[l]{\textbf{Ragas(96\%)} \\ Baseline(86\%) \\ Ranking(85\%)} & \shortstack[l]{Ragas(85\%) \\ Baseline(83\%) \\ \textbf{Ranking(90\%)}} & \shortstack[l]{\textbf{Ragas(79\%)} \\ Baseline(70\%) \\ Ranking(65\%)} \\
        \midrule
        \multicolumn{4}{l}{\textit{\textbf{English Dataset Evaluation}}} \\
        \midrule
        Gemini 3 Pro & \shortstack[l]{\textbf{Ragas(94\%)} \\ Baseline(90\%) \\ Ranking(92\%)} & \shortstack[l]{Ragas(84\%) \\ Baseline(89\%) \\ \textbf{Ranking(92\%)}} & \shortstack[l]{\textbf{Ragas(100\%)} \\ Baseline(90\%) \\ Ranking(90\%)} \\
        \midrule
        Gemini 3 Flash & \shortstack[l]{\textbf{Ragas(98\%)} \\ Baseline(86\%) \\ Ranking(85\%)} & \shortstack[l]{\textbf{Ragas(84\%)} \\ Baseline(70\%) \\ Ranking(82\%)} & \shortstack[l]{Ragas(8\%) \\ Baseline(20\%) \\ \textbf{Ranking(32\%)}} \\
        \midrule
        Gemini 2.5 Pro & \shortstack[l]{\textbf{Ragas(98\%)} \\ Baseline(85\%) \\ Ranking(88\%)} & \shortstack[l]{Ragas(68\%) \\ Baseline(69\%) \\ \textbf{Ranking(73\%)}} & \shortstack[l]{\textbf{Ragas(90\%)} \\ Baseline(85\%) \\ Ranking(78\%)} \\
        \bottomrule
    \end{tabular}
\end{table*}

\section{Results}

Table \ref{tab:evaluation_results} presents a comparison of the alignment between automated metrics and human judgments across three dimensions: Faithfulness, Answer Relevancy, and Context Relevancy. The evaluation contrasts the performance of three distinct judge models (Gemini 3 Pro, Gemini 3 Flash, and Gemini 2.5 Pro) across both Romanian (Low-Resource) and English (High-Resource) datasets.

\subsection{Performance on Romanian Data}
In the context of the Romanian dataset, the \textbf{Gemini 2.5 Pro} model demonstrated superior performance in assessing \textbf{Faithfulness}, achieving a 96\% alignment with human annotators using the Ragas-style decomposition method. This outperformed Gemini 3 Pro (84\%) and Gemini 3 Flash (87\%) using the same methodology. For \textbf{Answer Relevancy}, the Comparative Ranking (LLM-as-Ranker) approach proved most effective. Gemini 2.5 Pro again led the benchmarks with a 90\% agreement rate, whereas Gemini 3 Pro and Flash achieved lower correlations of 80\% and 77\%, respectively.

In \textbf{Context Relevancy}, \textbf{Gemini 3 Pro} utilizing the Ragas framework achieved the highest score in the entire Romanian evaluation set (93\%), demonstrating a good capability in filtering noise. In contrast, the lightweight Gemini 3 Flash model failed to produce meaningful signal for this metric, scoring between 0\% and 5\% across all methodologies.

\subsection{Performance on English Data}
On the English control dataset, performance was universally higher. \textbf{Gemini 3 Pro} achieved near-perfect alignment (100\%) for Context Relevancy using the Ragas method. Similarly, Faithfulness scores were elevated, with Gemini 3 Flash reaching 98\% accuracy via the Ragas method, surpassing its performance on the Romanian equivalent by 11 percentage points. 

\section{Discussion}

The results illuminate the disparities between high-resource and low-resource evaluation pipelines, while also highlighting the nuance required in selecting the appropriate "Judge" model.

A primary observation from Table \ref{tab:evaluation_results} is the consistent performance degradation when moving from English to Romanian. For instance, while Gemini 3 Pro achieved 94\% alignment for Faithfulness in English, this dropped to 84\% for Romanian. Similarly, Gemini 3 Flash's ability to assess Answer Relevancy dropped from 84\% (English) to 76\% (Romanian). This confirms our hypothesis that "LLM-as-a-Judge" systems are not language-agnostic, their reliability as evaluators is intrinsically tied to their pre-training exposure to the target language. Consequently, RAG pipelines for LRLs cannot blindly rely on default English-centric prompts and thresholds.

Our data suggests a bifurcation in the optimal evaluation strategy depending on the metric type. For objective metrics like \textbf{Faithfulness}, the granular Ragas-style decomposition consistently yielded the highest scores (e.g., 96\% for Gemini 2.5 Pro). This suggests that breaking down claims into atomic statements helps the model overcome linguistic ambiguity in Romanian. Conversely, for \textbf{Answer Relevancy}, the Comparative Ranking method frequently outperformed the Ragas decomposition (e.g., Gemini 2.5 Pro scored 90\% with Ranking vs. 85\% with Ragas). This indicates that for nuanced, holistic judgments of "quality", simpler pairwise comparisons are easier for the model to process than complex embedding-based synthetic generation.

Unexpectedly, the older \textbf{Gemini 2.5 Pro} architecture outperformed the newer Gemini 3 variants in several Romanian benchmarks, particularly Faithfulness (96\% vs. 84/87\%). While Gemini 3 Pro demonstrated superior reasoning in Context Relevancy (93\%), Gemini 2.5 Pro offered the most consistent all-around performance for the Romanian language. 

Furthermore, the failure of \textbf{Gemini 3 Flash} in Context Relevancy (scoring near 0\%) serves as a cautionary tale. While "Flash" models are attractive for their speed and cost, they appear to lack the depth of reasoning required to discern subtle noise in retrieved Romanian contexts, making them unsuitable as judges for this specific metric.

\section{Conclusion and Future Work}

This study bridges a gap in the evaluation of RAG systems for Low-Resource Languages. By introducing \textbf{AdminRo-Eval} and benchmarking three automated evaluation strategies, we demonstrated that English-centric metrics cannot be blindly transferred to Romanian without performance degradation.

Our experiments reveal that the "LLM-as-a-Judge" paradigm is viable for Romanian, provided the methodology is tailored to the specific metric. We found that the granular decomposition of the Ragas framework is superior for objective metrics like \textbf{Faithfulness} (96\% alignment with Gemini 2.5 Pro), whereas Comparative Ranking is more effective for nuanced semantic judgments like \textbf{Answer Relevance} (90\% alignment). Furthermore, our results caution against the use of lightweight "Flash" models for complex reasoning tasks in LRLs, as evidenced by their inability to accurately assess Context Relevance. Ultimately, while newer models like Gemini 3 Pro excel in specific areas, the older Gemini 2.5 Pro architecture provided the most robust and consistent baseline for Romanian RAG evaluation.

To further advance RAG evaluation for under-represented languages, we propose to expand the AdminRo-Eval dataset beyond public administration to include medical and legal domains, where hallucination risks carry higher stakes. Also, reliance on proprietary models (Gemini/GPT) limits accessibility for the LRL community, so we would want to assess the efficacy of open-weights models (e.g., Llama 3, Mistral) as judges also. 

\section{Limitations}

Our study faces three primary limitations. First, the \textbf{AdminRo-Eval} dataset focuses exclusively on formal public administration documents. Findings may not generalize to open-domain or informal contexts where semantic ambiguity is higher. Second, the modest size of our manually annotated ground truth (100 samples) limits the statistical power. Third, since Gemini was used to generate the synthetic test data, the high agreement scores may partially reflect a 'self-enhancement' bias, where the model preferentially favors its own generation style. 

\section{Ethics Statement}

This work utilizes the \textbf{AdminRo-Eval} dataset, which was constructed exclusively from public documents available on official Romanian government websites. We performed manual verification to ensure that no sensitive Personally Identifiable Information (PII) of private citizens was included in the corpus. Regarding the human evaluation, the native speaker annotators were volunteers. While our research aims to enhance the factual reliability of RAG systems, we acknowledge that the "Judge" models (Gemini) may still exhibit biases inherent in their pre-training data. We emphasize that automated evaluation tools for the public sector are intended to assist human decision-making, not to replace human oversight in administrative processes.

\section{Acknowledgments}

This work was supported by a grant of the Ministry of Research, Innovation and Digitization, CCCDI - UEFISCDI, project number PN-IV-P7-7.1-PTE-2024-0046, within PNCDI IV.

\bibliography{custom}

\begin{thebibliography}{13}
\providecommand{\natexlab}[1]{#1}

\bibitem[{Anghel et~al.(2025)Anghel, Petru-Theodor, Creanga, and
  Nisioi}]{anghel2025rals}
Florentina Anghel, Craciun Petru-Theodor, Claudiu Creanga, and Sergiu Nisioi.
  2025.
\newblock Rals: Resources and baselines for romanian automatic lexical
  simplification.
\newblock In \emph{Proceedings of the 2025 Conference on Empirical Methods in
  Natural Language Processing}.

\bibitem[{Es et~al.(2023)Es, James, Espinosa-Anke, and
  Schockaert}]{es2023ragas}
Shahul Es, Jithin James, Luis Espinosa-Anke, and Steven Schockaert. 2023.
\newblock \href {https://arxiv.org/abs/2309.15217} {Ragas: Automated evaluation
  of retrieval augmented generation}.
\newblock \emph{arXiv preprint arXiv:2309.15217}.

\bibitem[{Gemini~Team(2025)}]{geminiteam2023gemini}
Google Gemini~Team. 2025.
\newblock \href {https://arxiv.org/abs/2507.06261} {Gemini 2.5: Pushing the
  frontier with advanced reasoning, multimodality, long context, and next
  generation agentic capabilities}.
\newblock \emph{arXiv preprint arXiv:2507.06261}.

\bibitem[{{Gemini Team, Google DeepMind}(2025)}]{geminiteam2025gemini3}
{Gemini Team, Google DeepMind}. 2025.
\newblock \href
  {https://storage.googleapis.com/deepmind-media/Model-Cards/Gemini-3-Pro-Model-Card.pdf}
  {Gemini 3 technical report}.
\newblock Technical report, Google DeepMind.
\newblock Released November 18, 2025.

\bibitem[{Joshi et~al.(2020)Joshi, Santy, Budhiraja, Bali, and
  Choudhury}]{JoshiEtAl2020}
Pratik Joshi, Sebastin Santy, Amar Budhiraja, Kalika Bali, and Monojit
  Choudhury. 2020.
\newblock \href {https://doi.org/10.18653/v1/2020.acl-main.560} {The state and
  fate of linguistic diversity and inclusion in the {NLP} world}.
\newblock In \emph{Proceedings of the 58th Annual Meeting of the Association
  for Computational Linguistics}, pages 6282--6293.

\bibitem[{Kreutzer et~al.(2022)Kreutzer, Caswell, Wang, Wahab, van Esch
  et~al.}]{KreutzerEtAl2022}
Julia Kreutzer, Isaac Caswell, Lisa Wang, Ahsan Wahab, Daan van Esch, and 1
  others. 2022.
\newblock \href {https://doi.org/10.1162/tacl_a_00447} {Quality at a glance: An
  audit of web-crawled multilingual datasets}.
\newblock \emph{Transactions of the Association for Computational Linguistics},
  10:50--72.

\bibitem[{Lin(2004)}]{lin-2004-rouge}
Chin-Yew Lin. 2004.
\newblock \href {https://aclanthology.org/W04-1013} {{ROUGE}: A package for
  automatic evaluation of summaries}.
\newblock In \emph{Text Summarization Branches Out: Proceedings of the ACL-04
  Workshop}, pages 74--81, Barcelona, Spain. Association for Computational
  Linguistics.

\bibitem[{Marchitan et~al.(2024)Marchitan, Creanga, and
  Dinu}]{marchitan2024transformer}
Teodor-George Marchitan, Claudiu Creanga, and Liviu~P Dinu. 2024.
\newblock Transformer and hybrid deep learning based models for
  machine-generated text detection.
\newblock In \emph{Proceedings of the 18th International Workshop on Semantic
  Evaluation (SemEval-2024)}, pages 403--411.

\bibitem[{Min et~al.(2023)Min, Krishna, Lyu, Lewis, Yih, Koh, Iyyer,
  Zettlemoyer, and Hajishirzi}]{min2023factscore}
Sewon Min, Kalpesh Krishna, Xinxi Lyu, Mike Lewis, Wen-tau Yih, Pang~Wei Koh,
  Mohit Iyyer, Luke Zettlemoyer, and Hannaneh Hajishirzi. 2023.
\newblock \href {https://arxiv.org/abs/2305.14251} {{FActScore}: Fine-grained
  atomic evaluation of factual precision in long form text generation}.
\newblock In \emph{Proceedings of the 2023 Conference on Empirical Methods in
  Natural Language Processing (EMNLP)}, pages 12076--12100.

\bibitem[{Ouyang et~al.(2022)Ouyang, Wu, Jiang, Almeida, Wainwright, Mishkin,
  Zhang, Agarwal, Slama, Ray et~al.}]{ouyang2022training}
Long Ouyang, Jeff Wu, Xu~Jiang, Diogo Almeida, Carroll~L Wainwright, Pamela
  Mishkin, Chong Zhang, Sandhini Agarwal, Katarina Slama, Alex Ray, and 1
  others. 2022.
\newblock Training language models to follow instructions with human feedback.
\newblock In \emph{Advances in Neural Information Processing Systems},
  volume~35, pages 27730--27744.

\bibitem[{Papineni et~al.(2002)Papineni, Roukos, Ward, and
  Zhu}]{papineni-etal-2002-bleu}
Kishore Papineni, Salim Roukos, Todd Ward, and Wei-Jing Zhu. 2002.
\newblock \href {https://aclanthology.org/P02-1040} {{B}leu: a method for
  automatic evaluation of machine translation}.
\newblock In \emph{Proceedings of the 40th Annual Meeting of the Association
  for Computational Linguistics}, pages 311--318, Philadelphia, Pennsylvania,
  USA. Association for Computational Linguistics.

\bibitem[{Zhang et~al.(2020)Zhang, Kishore, Wu, Weinberger, and
  Artzi}]{zhang2020bertscore}
Tianyi Zhang, Varsha Kishore, Felix Wu, Kilian~Q Weinberger, and Yoav Artzi.
  2020.
\newblock Bertscore: Evaluating text generation with bert.
\newblock In \emph{International Conference on Learning Representations}.

\bibitem[{Zheng et~al.(2023)Zheng, Chiang, Sheng, Zhuang, Wu, Zhuang, Lin, Li,
  Li, and P.~Stoica}]{zheng2023judging}
Lianmin Zheng, Wei-Lin Chiang, Ying Sheng, Siyuan Zhuang, Zhanghao Wu, Yonghao
  Zhuang, Zi~Lin, Zhuohan Li, Dacheng Li, and Ion P.~Stoica. 2023.
\newblock \href
  {https://papers.nips.cc/paper_files/paper/2023/hash/91f18a1287b398d378ef22505bf41832-Abstract-Datasets_and_Benchmarks.html}
  {Judging {LLM}-as-a-judge with {MT-Bench} and {Chatbot Arena}}.
\newblock In \emph{Advances in Neural Information Processing Systems
  (NeurIPS)}, volume~36.

\end{thebibliography}

\clearpage
\appendix
\onecolumn 

\section{Annotation Pipeline}
\label{sec:appendix}

The following figures illustrate the annotation interfaces used for human evaluation of the three metrics.

\begin{figure}[h!]
    \centering
    \includegraphics[width=0.7\linewidth]{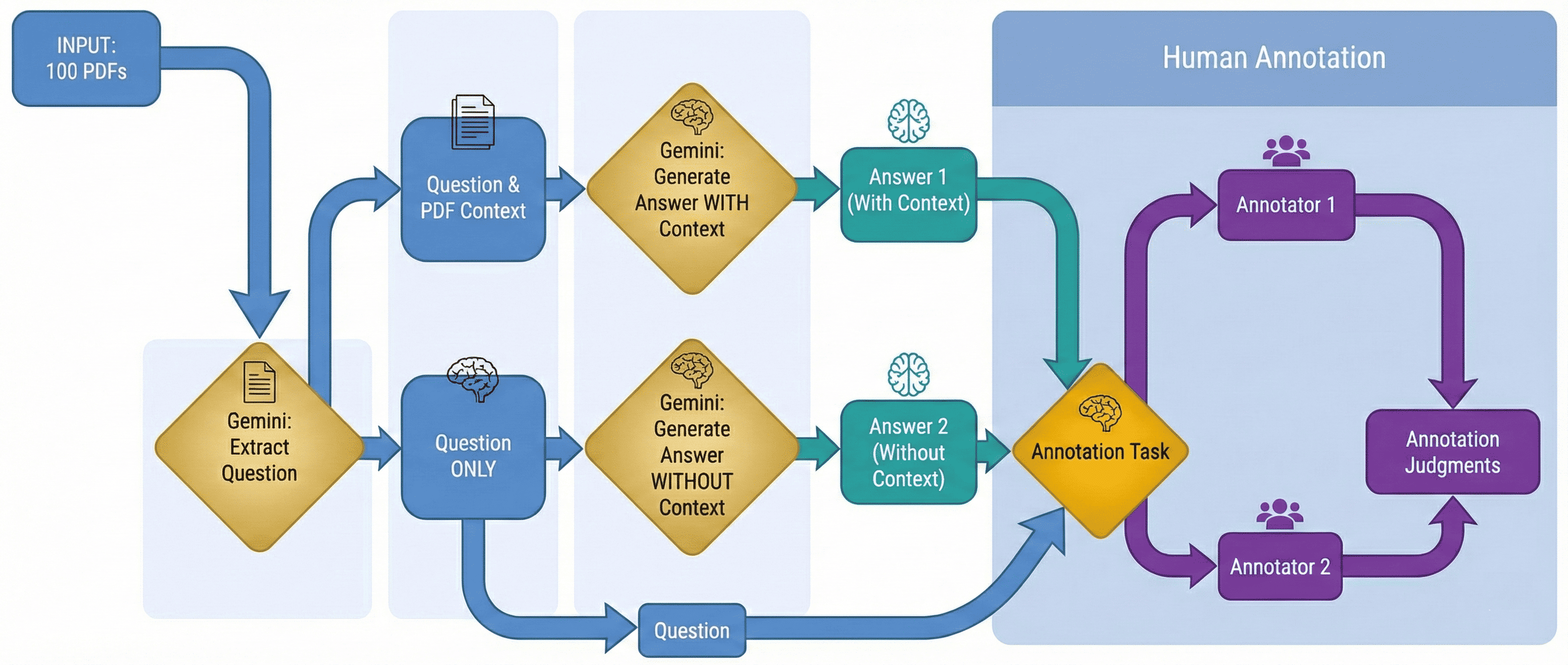}
    \caption{Interface for Faithfulness annotation. Reviewers examined the question, context, and two candidate answers to determine which response was strictly supported by the source text.}
    \label{fig:annotation_faithfulness}
\end{figure}

\begin{figure}[h!]
    \centering
    \includegraphics[width=0.7\linewidth]{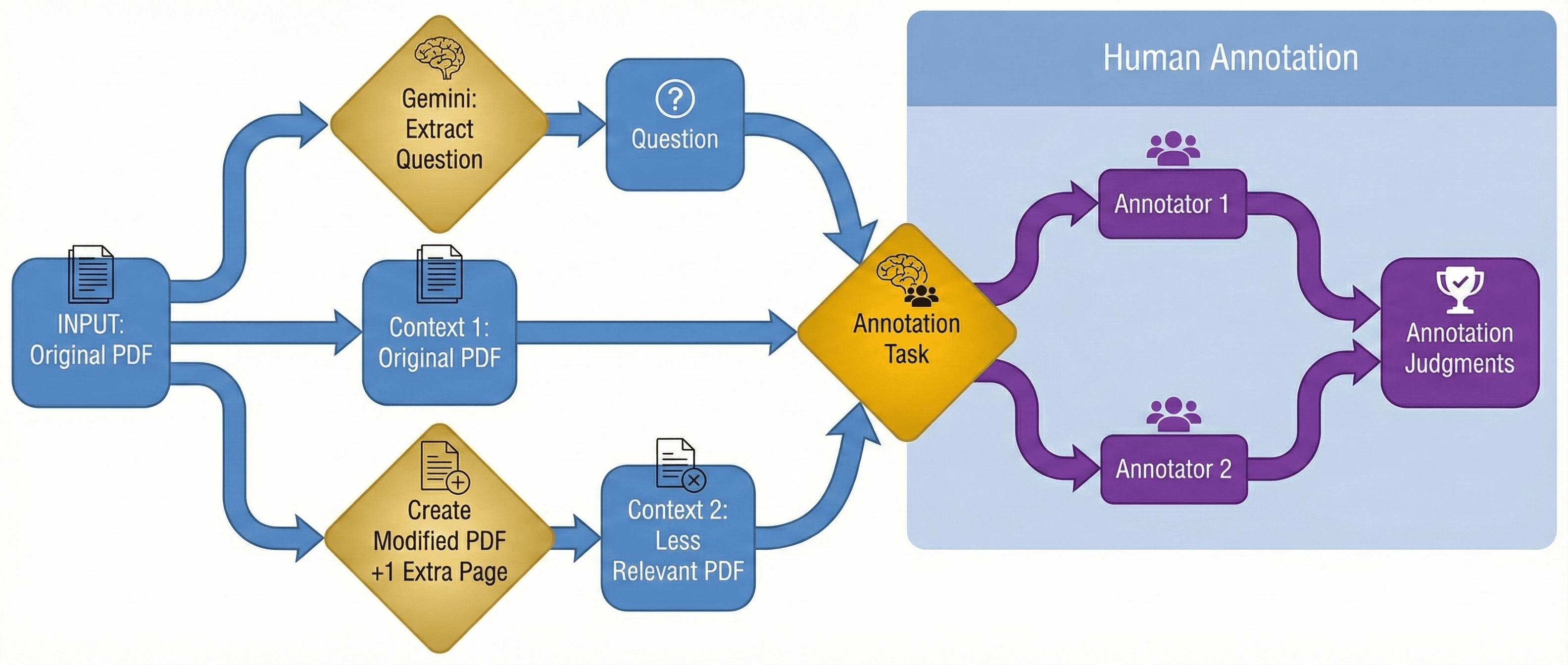}
    \caption{Interface for Context Relevance annotation. Annotators selected the most useful passage for answering the prompt, choosing between the original context and a version modified with extraneous text.}
    \label{fig:annotation_context}
\end{figure}

\begin{figure}[h!]
    \centering
    \includegraphics[width=0.7\linewidth]{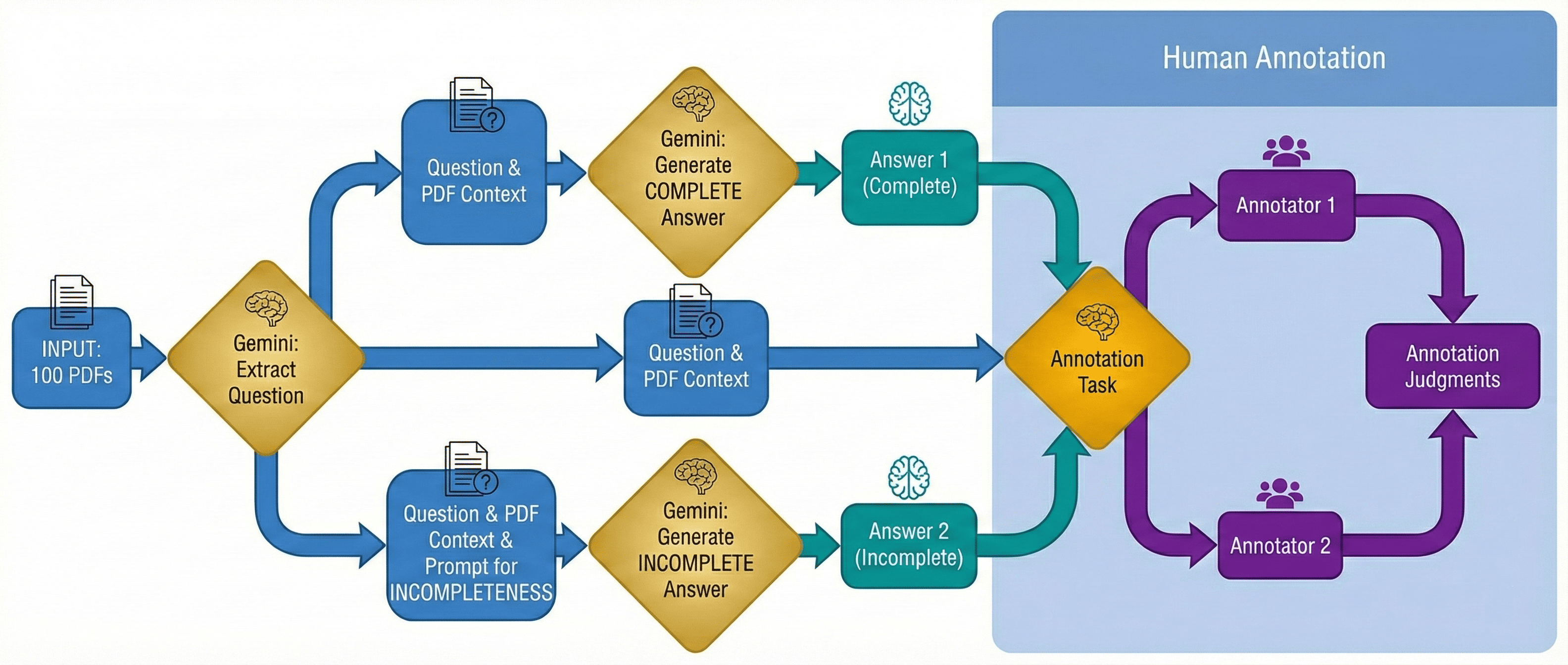}
    \caption{Interface for Answer Relevance annotation. The task required selecting the most helpful response by comparing a complete, direct answer against an intentionally partial or evasive one.}
    \label{fig:annotation_answer}
\end{figure}

\end{document}